\documentclass{article}

\usepackage{graphicx} % more modern
\usepackage{color}
\usepackage{algorithm}
\usepackage{algorithmic}
\usepackage{amssymb}
\usepackage{amsthm}
\usepackage{amsmath}
\usepackage{multirow}
\usepackage{threeparttable}
\usepackage{icml2014}
\icmltitlerunning{Learning Hidden Structures with Relational Models by Adequately Involving Rich Information in A Network}
\begin{document}

\twocolumn[
\icmltitle{Learning Hidden Structures with Relational Models \\by Adequately Involving Rich Information in A Network}

% It is OKAY to include author information, even for blind
% submissions: the style file will automatically remove it for you
% unless you've provided the [accepted] option to the icml2013
% package.
\icmlauthor{Your Name}{email@yourdomain.edu}
\icmladdress{Your Fantastic Institute,
            314159 Pi St., Palo Alto, CA 94306 USA}
\icmlauthor{Your CoAuthor's Name}{email@coauthordomain.edu}
\icmladdress{Their Fantastic Institute,
            27182 Exp St., Toronto, ON M6H 2T1 CANADA}

% You may provide any keywords that you
% find helpful for describing your paper; these are used to populate
% the "keywords" metadata in the PDF but will not be shown in the document
\icmlkeywords{boring formatting information, machine learning, ICML}
]
\vskip 0.3in

\begin{abstract}
Effectively modelling hidden structures in a network is very practical but theoretically challenging. Existing relational models only involve very limited information, namely the binary directional link data, embedded in a network to learn hidden networking structures. There is other rich and meaningful information (e.g., various attributes of entities and more granular information than binary elements such as ``like'' or ``dislike'') missed, which play a critical role in forming and understanding relations in a network. In this work, we propose an \emph{informative relational model} (InfRM) framework to adequately involve rich information and its granularity in a network, including \emph{metadata information} about each entity and various forms of \emph{link data}. Firstly, an effective metadata information incorporation method is employed on the prior information from relational models {MMSB} and {LFRM}. This is to encourage the entities with similar metadata information to have similar hidden structures. Secondly, we propose various solutions to cater for alternative forms of link data. Substantial efforts have been made towards modelling appropriateness and efficiency, for example, using conjugate priors. We evaluate our framework and its inference algorithms in different datasets, which shows the generality and effectiveness of our models in capturing implicit structures in networks.
\end{abstract}

\section{Introduction}
%Relational learning in social networks has become a popular research topic as the large need in digging the social structure. One promising aspect of it is that it can be directly applied to other fields, such as protein-to-protein interaction\cite{}, social media recommendation\cite{}, and food-chain discovery\cite{}. By placing a proper generative process on the link data, the learning results provide interesting insights into the understanding of social structure. Correspondingly, these work is defined as {relational models}, with benchmark examples listed as {infinite relational model (IRM), latent feature relational model (LFRM) and mixed membership stochastic blockmodel (MMSB)}.
%
%The Latent Feature Relational Model (LFRM) \cite{miller2009nonparametric} has gained success in predicting the missing link. However, the latent feature matrix $Z$ discovered by {LFRM} did not provide an illustrative understanding of the the hidden structure, since the metadata information is incorporated in the likelihood generation. Other benchmark model in relational modeling includes the Infinite Relational Model (IRM) \cite{kemp2006learning} and Mixed-Membership Stochastic Blockmodel (MMSB) \cite{airoldi2008mixed}. Also, there is work \cite{PalKnoGha12} in discovering a hierarchical structure of the latent features.
Learning hidden structures within a network is an emergent topic in various areas including social-media recommendation \cite{tang2010community}, customer partitioning, social network analysis, and partitioning protein interaction networks \cite{girvan2002community, fortunato2010community}. Many models have been proposed in recent years to address this problem by using linkage information such as a person's view towards others. Examples include \emph{stochastic blockmodel} \cite{doi:10.1198/016214501753208735} and its infinite communities case \emph{infinite relational model} (IRM) \cite{kemp2006learning}, both aiming at partitioning a network of entities into different groups based on their pairwise, directional binary observations. The ``inter-nodes'' link data used in existing approaches contributes to the explicit insight on social structures.

On the other hand, the ``intra-nodes'' metadata information could complement the disclosure of hidden implicit relations structures. Let us take the Lazega lawfirm network (detailed in Section \ref{sec_5}) as an example, which contains both link and metadata information: The \emph{metadata information} include attributes such as offices (Boston, Hartford or Providence), law schools (harvard, yale, ucon or other) and age associated with each of the entities (attorneys). Naturally, people with similar attributes tend to have relationship with each other. The directional \emph{link data} including elements such as basic advice frequency, co-work time and friendship. These elements however, may take different forms. Here the first two elements can be represented by integer values (i.e. count link data), while friendship may better be represented by a real value on the unit interval (i.e. unit link data). If they have to be ``binarized'' as in the case of the existing models, then the lost of granularity could lead to poorer utilization of information.
%
%Count link data, e.g., the communication frequencies, the degree of like or dislike and the co-work times, and unit link data, e.g. the proportion of agreement between the two points of view,

While some of the most recent efforts are directed to involve more information, they all face with some shortcomings. For instance, in LFRM \cite{miller2009nonparametric}, to generate the link data, a combination of metadata information with the feature matrix introduces the ambiguity into the role of latent feature matrix. In terms of efficient inference, as shown in NMDR \cite{conf/icml/KimHS12}, the logistic-normal transform was employed to integrate the metadata information into each entity's membership distribution. However, this integration complicates the original structures and leads to a non-conjugacy. In terms of link-data modelling, in order to use count link data as observations, \cite{Yangml11} has used geometric distribution as its likelihood model. This implies the probabilities of counts decreases monotonically, which may not be applicable in a general setting.

%However, their work is restricted in {MMSB} and more importantly, it suffers the non-conjugacy problem while inferencing the latent communities. This is vital to both of the MCMC inference and variational inference method.

%All the above relational models are using one informative binary linkage data, while information is provided more than the linkage data, e.g., the entity's attribute, an count variable indicate the closeness of a relationship, it would be a promising direction in exploring how to effectively using this metadata information. More frequency of basic advices seeking indicates closer relationship, rather than a simplified ``existed'' or ``non-existed''. Also, two entities pair obsess all the above three relations would be closer than the one with one single relation. Aside from the binary version in link data, other forms exist in real applications, which are incompatible with the default setting of the previous relational models. Thus, proper emission distributions are needed for the individual modelling.
%
%mentioned slightly on modelling the and claims to use a geometric distribution as the emission distribution for modelling. However, the assumption of larger frequency data's probability is smaller than the smaller one' in geometric distribution violates the real world rules and there is also no experimental validation on this.
%\cite{ren2011logistic} use a logistic stick breaking way to construct the bayesian nonparametric prior. \cite{rodriguez2011nonparametric} use the probit stick breaking method to finish the bayesian nonparametric prior setting.

We propose a new \emph{informative relational model} (InfRM) framework, which incorporates both the rich metadata information and granular link data in a sensible and efficient way. To integrate the metadata information, a new transform is proposed, and the corresponding result is placed as the prior for the communities of each entity. This enables the similarity of metadata information between entities to be reflected in their corresponding hidden structures. Inspired by the existing benchmark relational models of MMSB \cite{airoldi2008mixed} and LFRM \cite{miller2009nonparametric}, we individually model the hidden structure as mixed memberships and latent features respectively. This lead to \emph{informative mixed membership model} (InfMM) and \emph{informative latent feature model} (InfLF) which is the centerpiece of this paper. The stick breaking process \cite{citeulike:1495491}\cite{TehGorGha2007} alike methods were proposed to model the unknown number of communities. In particular, our InfMM model successfully gains the conjugate property. As discussed in Section \ref{sec_31}, through these efforts, the existing models can be seen as the special cases of our proposed models.
%This retains some elegant properties of the Dirichlet Process and more importantly, enjoys the conjugate property on inferencing both of the mixed-membership distributions and the metadata importance indicator.

In addition, we designed a set of solutions to model the various forms of link data, including the count and unit link data. We have chosen the likelihood and prior model carefully for both of their practical appropriateness and computation efficiencies. Their effects have been demonstrated in our experiments.

%Thus, we summarize our work's contributions into three parts. Firstly, we have successfully incorporated the metadata information into the latent feature matrix and more importantly, produced a comprehensive result. Secondly, we successfully change the transform function and make it conjugate without loss of the original property. Last, we have considered integer or even real links' value, more than only the binary ones, this enables us to fully utilize the given information.

As a result, our models capture much richer information embedded in a network, thus leading to better performance as illustrated in Section \ref{sec_5} in modelling hidden structures. The rest of the paper is organized as follows. Section \ref{sec_3} introduces the relational models and necessary notations for our work. In Section \ref{sec_2}, we describe the InfRM framework for integrating metadata information into each entity's hidden structure, including  InfMM and InfLF. The generation distributions proposed for different forms of link data are provided in Section \ref{sec_link_data}. Section \ref{sec_4} discusses the sampling methods and computational complexity analysis. Experiments in Section \ref{sec_5} compare our methods with the previous work and validate our model performance. Conclusions and future works are in Section \ref{sec_6}.

\section{Relational Models \& Notations \label{sec_3} }
\subsection{Relational Models}
The  \emph{stochastic blockmodel} \cite{doi:10.1198/016214501753208735} assumes that each entity has a latent variable that directly represents its community membership. Each of the fixed number of communities associates with a weight, and the whole weight vector can be seen as a draw from a $K$-dimensional Dirichlet distribution. Naturally, the community memberships are realized from the multinomial distribution parameterized by the weight vector. The binary link data between two entities is determined by their belongingness communities. This model has been extended to an infinite $K$ community, i.e., \emph{infinite relational model} (IRM) \cite{kemp2006learning} where the Dirichlet distribution has been replaced by a Dirichlet process.

Various recent work has been proposed to capture the complex interactions amongst entities based on \emph{stochastic blockmodel}, which can be categorized into two notable branches, both are a generalisation of \emph{stochastic blockmodel}. The \emph{first branch} features the \emph{latent feature relational model} (LFRM) \cite{miller2009nonparametric}: instead of associating an entity with only a single feature, i.e., its membership indicator, it allows a variable number of binary features to be associated with each entity. The \emph{second branch} follows the \emph{mixed-membership stochastic blockmodel} (MMSB) model, in which each entity has its own community distribution, hence having a ``mixed'' class of interactions with other entities.

The LFRM-like work was originated from \cite{hoff2002latent,hoff2005bilinear}, while it assumes a latent real-valued feature vector for each entity. The LFRM in \cite{miller2009nonparametric} uses a binary vector to represent latent features of each entity, and the number of features of all entities can potentially be infinite using an Indian Buffet Process prior \cite{citeulike:6102284, griffiths2011indian}. The work in \cite{PalKnoGha12} further uncovers the substructure within each feature and uses the ``co-active'' features from two entities during generating their link data. On the MMSB-typed work, a few variants have been subsequently proposed, including \cite{koutsourelakis2008finding} which extends the MMSB into the infinite community case and \cite{doi:10.1080/01621459.2012.682530} which uses the nested Chinese Restaurant Process \cite{Blei:2010:NCR:1667053.1667056} to build the  hierarchical structure of communities.

\subsection{Notations}
%We here give an extensive explanation of the variables as our model would inherit this notation system. $\phi_i$ denotes the metadata information we currently have. $\phi$ is expressed as an $N\times F$ binary matrix $\Phi$, where $\Phi_{if}=1$ denotes the $i^{\textrm{th}}$ data occupies the $f^{\textrm{th}}$ character, otherwise $\Phi_{if}=0$. Correspondingly, there is an $F\times K$ matrix $\eta$ with $\eta_{fk}$ indicating the importance of $f^{\textrm{th}}$ character to $k^{\textrm{th}}$ roles. We should note that in {NMDR}, this could either be a positive factor or a negative one. The relational data $e$ is an $N\times N$ asymmetric binary matrix, with $e_{ij}=1$ denoting a relationship exists from entity $i$ to entity $j$, otherwise $e_{ij}=0$.
All notations in this paper are given in Table \ref{table_3}.
\begin{table}[htbp]
\caption{Notations for our {InfRM}} \label{table_3}
\centering
\begin{tabular}{c|l}
\hline
$n$ & number of entities\\
  \hline
$K$ & number of discovered communities\\
  \hline
$F$ & number of attributes in metadata\\
  \hline
\multirow{2}{*}{$\phi$} & an $n\times F$ binary matrix, $\phi_{if}=1$ denotes\\
& the $i^{\textrm{th}}$ data occupies the $f^{\textrm{th}}$ attribute\\
\hline
\multirow{2}{*}{$\eta$} & an $F\times K$ positive matrix, $\eta_{fk}$ indicates\\
& the importance of $f^{\textrm{th}}$ attribute to $k^{\textrm{th}}$ roles.\\
\hline
$e_{ij}$ & directional, binary interactions\\
\hline
${\small s_{ij}, r_{ij}}$ & membership indicators of $e_{ij}$ in MMSB\\
\hline
$z_i$ & latent feature vector of entity $i$ in LFRM\\
  \hline
\multirow{2}{*}{$\pi_i$} & membership distribution for entity $i$, $\pi_{ik}$ is\\
 & the significance of community $k$ for  entity $i$ \\
   \hline
\multirow{2}{*}{$B$} & asymmetric, role-compatibility matrix, $B_{kl}$\\
 & indicates compatibility of communities $k, l$ \\
  \hline
  \end{tabular}
\end{table}

\section{Involving Metadata Information \label{sec_2}}
%\begin{figure}
%\centering
%\includegraphics[scale=0.35, width = 0.35 \textwidth, bb = 148 528 288 671, clip]{images/cncrp_frame.eps}
%\caption{Graphical model of Informative Infinite Roles Relational Models}
%\label{fig_1}
%\end{figure}

\begin{figure}[htbp]
\centering
\includegraphics[scale=0.5, width= 0.35 \textwidth, bb = 147 507 277 674, clip]{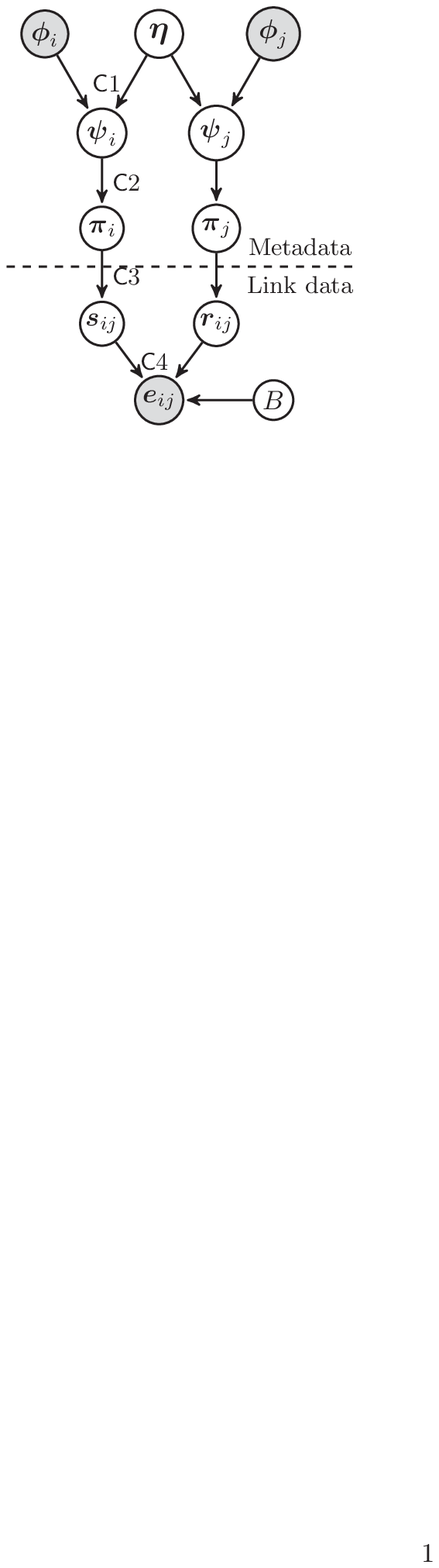}
\caption{The generative model for several implementations. Observations are denoted in grey. The part above the dashed line corresponds to involving metadata information in Section 3; the below part corresponds to modeling link information discussed in Section 4. C$1$ to C$4$ represent four conditional distributions in two different forms as shown in Sections \ref{sec_31} and \ref{sec_32}, respectively.}\label{fig_1}
\end{figure}

Figure \ref{fig_1} depicts the generative models of all the variables used in our work. In this paper, metadata information is incorporated into both branches of the stochastic blockmodel described earlier, i.e. MMSB and LFRM. Further, it can be applied to their base mode IRM \cite{kemp2006learning} with the similar approach, which we will not elaborate here.

%
%In integrating these metadata information, we proposed two models here, first with the {logistic-normal latent feature relational model (LN-LFRM)} extends the {NMDR} into the {LFRM} case and then comes the {informative mixed membership model (InfMM)} incorporates the metadata information into each communities' generative parameter while at the same time enables the importance indicator matrix to be conjugately inferred.
%%
%Assuming we have obtained the importance indicator matrix $\eta$, then the problem becomes as combining the Character matrix $\Phi$ and the $\eta$ together to signify the prior probability of each role. Two methods, i.e., logistic normal and beta distribution, are proposed and discussed to ``transfer'' these two variables into the role probability $\{\psi_{ik}\}_{i,k}$.

\subsection{Informative Mixed Membership Model \label{sec_31}}
The generative process for \emph{informative mixed membership} (InfMM) model is defined as follows (W.l.o.g. $\forall i,j=1,\ldots,n,k\in N^+$):
\begin{description}
\item{C$1$}, $ \psi_{ik}\sim \textrm{Beta}(1, \prod_f \eta_{fk}^{\phi_{if}});$
\item{C$2$}, $ \pi_{ik}=\psi_{ik}\prod_{l=1}^{k-1}(1-\psi_{il}); $
\item{C$3$}, $ s_{ij}\sim \textrm{Multi}(\pi_i), r_{ij}\sim \textrm{Multi}(\pi_j); $
\item{C$4$}, $ e_{ij}\sim g(B_{s_{ij}r_{ij}}).$
\end{description}
Here C$1$ and C$2$ are the stick breaking representation for our mixed membership distribution $\pi_i$. C$3$ and C$4$ correspond to the membership indicator and link data's generation, respectively. Detailed elaboration is in \cite{airoldi2008mixed,koutsourelakis2008finding}. We leave equation C$4$ in its general form, i.e., $g(B_{s_{ij}r_{ij}})$, which  may take on a variety of forms, such as those described in Section \ref{sec_link_data}.

We use the attribute {\it age} in Lazega lawfirm to further explain the importance indicator $\eta_{fk}$ used in C$1$. W.l.o.g., we let $f_0^{\text{th}}$ column of $\phi$ matrix denote the age attribute of all the entities,  $\phi_{if_0}=1$ implies that entity $i$ has $age > 40$ (in our experimental setting), and $ 0 $ otherwise. From equation C$1$, one can easily see that when $\eta_{f_0k}\ll 1$, {\it age} would largely increase the impact of the $k^{\textrm{th}}$ community. Likewise, $\eta_{f_0k}\gg 1$ reduces the significance of the {\it age} attribute on the $k^{\textrm{th}}$ community. $\eta_{f_0k}=1$ means that {\it age} does not have influence on the $k^{\textrm{th}}$ community at all. When $\phi_{if_0}=0$, it makes {\it age} of the entity $i$ neutral towards all other communities.

%On the explanation of importance indicator $\eta_{fk}$ in C$1$, we take the attribute \emph{Office} in Lazega lawfirm as an instance. in our {InfMM}, smaller value of attribute $f$ would promote the popularity of community $k$. Take the attribute {office} in Lazega lawfirm as an instance, smaller value indicates it account
%
%This is to be further explained in the experiments.
%
%a smaller $\eta_{fk}\in R$ value in {NMDR} indicates less importance of this attribute $f$ for community $k$; while in our {InfMM}, $\eta_{fk}\in R^+$ and and larger value prohibit it, as we treat the product as the second parameter of the beta distribution.

Instead of C$1$, the NMDR model \cite{conf/icml/KimHS12,kim2011doubly} uses the logistic normal distribution (with the mean value being the linear sum (i.e., $\sum_{f}\phi_{if}\eta_{fk}$)) to construct a stick-breaking weight $\psi_{ik}$. While the method can successfully integrate the metadata information into the entity's membership distribution, it suffers from the lack of conjugacy, which makes inference inefficient. In our approach InfMM, we replaced the logistic normal distribution with a beta distribution, parameterised by $\prod_f \eta_{fk}^{\phi_{if}}$, where the positive, importance indicator $\eta_{fk}$ is given a vague gamma prior $\eta_{fk}\sim Gamma(\alpha_{\eta}, \beta_{\eta})$.

%In this second target, we are still targeting incorporating the metadata information into the components' weights, however, differs in introducing into the parameter of beta distribution in Eq. (\ref{eq_5}), with an {exponential-sum-logarithm (ESL)} construction.
%\begin{equation}
%\begin{aligned}
%;\\
%\psi_{ik}\sim Beta(1, \prod_f\eta_{fk}^{\phi_{if}}).\\
%\end{aligned}
%\end{equation}

%Comparing to the conventional arithmetic mean (i.e., linear transform) $\sum_f \phi_{if}\eta_{fk}$, our {ESL} metadata information construction is the power of the sum of logarithm-transformed $\eta$, i.e., $\prod_f\eta_{fk}^{\phi_{if}}=\exp(\sum_{f}\phi_{if}\ln\eta_{fk})$. They can achieve similar results while presented in different forms.
%
%The $\eta_{fk}$ here denotes the ``importance indicator'' of feature $f$ for the $k^{\textrm{th}}$ component in the underlying completely random measure. While $\eta_{fk}<1$, the presence of this feature $f$ would downgrade the importance of component $k$; while $\eta_{fk}>1$, the corresponding component $k$'s importance would increase. While the metadata information $\phi_{if}=0$, it will keep neutral towards the importance of feature $f$ as $\eta_{fk}^{\phi_{if}}=1$, this is the same as previous work \cite{conf/icml/KimHS12}.
%
%%On the differences of {LN} and {ESL} from the variable transform perspective, our new construction can be regarded as replacing the logistic normal distribution with one beta process and the normal distribution with one gamma distribution.
This operation leads to a conjugate property we can enjoy, on both of the importance indicator $\eta_{fk}$ and stick-breaking weight $\psi_{ik}$. More specifically, the distributions of $\eta_{fk}, \psi_{ik}$ are:
{
\begin{equation}
\begin{split}
p(\eta_{fk}|\alpha_{\eta}, \beta_{\eta}) & \propto\eta_{fk}^{\alpha_{\eta}-1}e^{-\beta_{\eta}\eta_{fk}};\\
p(\psi_{ik}|\eta_{\cdot k}, \phi_i) & \propto\left[\prod_f\eta_{fk}^{\phi_{if}}\right]\cdot(1-\psi_{ik})^{\prod_f\eta_{fk}^{\phi_{if}}-1}.
\end{split}
\end{equation}
}
Thus, the posterior distribution of $\eta_{fk}$ becomes:
{\small
\begin{equation}
\eta_{fk}\sim \textrm{Gamma}(\alpha_{\eta}+\sum_i\phi_{if}, \beta_{\eta}-\sum_i\phi_{if}\ln(1-\psi_{ik})\prod_{F\neq f}\eta_{Fk}^{\phi_{iF}})
\end{equation}
}

The joint probability of $\{s_{ij},r_{ji}\}_{j=1}^n$ becomes:
\begin{equation}
p(\{s_{ij}\}_{j=1}^n,\{r_{ji}\}_{j=1}^n|\psi_{i\cdot})\propto\prod_{k=1}^K\left[\psi_{ik}^{N_{ik}}(1-\psi_{ik})^{\sum_{l=k+1}^K N_{il}}\right]
\end{equation}
here $N_{ik}=\#\{j:s_{\boldsymbol{i}j}=k\}+\#\{j:r_{j\boldsymbol{i}}=k\}$.

The posterior distribution of $\psi_{ik}$ becomes:
\begin{equation} \label{eq_1}
\psi_{ik}\sim \textrm{Beta}(N_{ik}+1, \sum_{l=k+1}^K N_{il}+\prod_f\eta_{fk}^{\phi_{if}})
\end{equation}
The posterior distribution in Eq. (\ref{eq_1}) is consistent with the result in \cite{ishwaran2001gibbs,kalli2011slice}, where their result is conditioned on single concentration parameter $\alpha$ instead of $\prod_f\eta_{fk}^{\phi_{if}}$.

%Another interesting comparison is the difference between the classic iMMM and our {InfMM}. In iMMM, the mixed membership distribution is a draw from the Dirichlet Process with fixed concentration parameter $\alpha$, however, in our {InfMM}, this constant concentration parameter $\alpha$ is treated unequally for different communities: with community $\pi_{ik}$ for parameter $\prod_f\eta_{fk}^{\phi_{if}}$. With this unequal parameter, we can achieve our goal of introducing the metadata information into the mixed membership distribution.

Another interesting comparison is the placing of prior information for communities within different models. In iMMM, although the author claimed to use different $\alpha_i$ to model individual $\pi_i$, however, each stick-breaking weight $\psi_{ik}$ within one $\pi_i$ is generated identically, i.e., from $\text{beta}(1, \alpha_i)$. This is still insufficient for many practical applications. Accordingly NMDR has incorporated metadata information using logistic normal function, as stated above. In a way, this approach has further generalised the model, such that each $\psi_{ik}$ differs in their distributions.

Despite the model relaxation, empirical results show that NMDR has a slow convergence. It is therefore imperative for us to search for a more efficient way to incorporate the metadata information. Compared to iMMM, our InfMM model replaces unified $\{\alpha_i\}$ with $\prod_f\eta_{fk}^{\phi_{if}}$ for the generation of $\psi_{ik}$. Its conjugate property makes our model appealing in terms of mixing efficiency, which is confirmed in the results shown in Section \ref{sec_61}. What is more is that our model can be seen as a natural extension of the popular iMMM model. By letting $\eta_{fk} = \alpha^{1/F}$ and $\phi_{if} = 1$ , we obtain the classical iMMM. This makes sense, as without the presence of metadata, each feature is assumed to be counted equally, which implies that the model becomes the classical iMMM.

\subsection{Informative Latent Feature Model \label{sec_32}}
%\begin{equation} \label{eq_5}
%\begin{aligned}
%{\eta}_{fk}\sim N({\mu}_f, {\lambda}_f^2), \forall k\ge1\\
%{v}_{ik}\sim N(\phi_{i\cdot}^T\cdot\eta_{\cdot k}, {\lambda}_v^2)\\
%\psi_{ik}=\frac{1}{1+\exp{(-{v}_{ik}})}.\\
%\pi_{ik}=\psi_{ik}\prod_{l=1}^{k-1}(1-\psi_{il})\\
%\end{aligned}
%\end{equation}
The generative process for \emph{informative latent feature} (InfLF) model is defined as follows:
\begin{description}
\item{C$1$}, $ \psi_{ik}\sim \textrm{Beta}(\prod_f \eta_{fk}^{\phi_{if}}, 1);$
\item{C$2$}, $ \pi_{ik}=\prod_{l=1}^{k}\psi_{il}; $
\item{C$3$}, $ z_{ik}\sim \textrm{Bernoulli}(\pi_{ik}); $
\item{C$4$}, $ e_{ij}\sim g(z_{i}Bz_j^T).$
\end{description}
Here C$1$ and C$2$ refer to the detailed construction of our specialized stick breaking representation $\pi_i$. Similar to the traditional stick-breaking process \cite{citeulike:1495491}\cite{TehGorGha2007}, they are used to generate the latent feature matrix $z$ in C$3$. C$4$ corresponds to the link data's generation in our model. Our work can be seen as an extension to the traditional LFRM, which can be seen at \cite{miller2009nonparametric}.

%\cite{thibaux2007hierarchical} with concentration parameter equals $1$. This is better known as Completely Random Measure (CRM) of the Indian Buffet Process. However, the () assumes that

However, our InfLF's hidden structure differs from the one of LFRM. More specifically, the original LFRM uses \emph{one} specialized beta process as the underlying representation for all the $n$ entities' latent feature $z$. This process can be easily marginalized out in convenient of the Beta-Bernoulli conjugacy \cite{thibaux2007hierarchical}. In our InfLF, each $i^{\textrm{th}}$ entity's latent feature is motivated by their own stick breaking representation $\pi_i$, i.e., there are $\boldsymbol{n}$ representations in total. Thus, the individual metadata information is contained in each corresponding representation, which will be reflected in the latent feature.

We use the new transform, i.e., $\prod_f \eta_{fk}^{\phi_{if}}$,  as the mass parameter \cite{thibaux2007hierarchical} in the construction of the stick breaking representation, as stated in C$1$. The importance indicator $\eta$ here plays an opposite role when comparing to the InfMM model, i.e., larger value of $\eta_{fk}$ would make the present of attribute $f$ promote the $k^{\textrm{th}}$ community.

An interesting notation is that the stick breaking representations in both of our InfMM and InfLF are no longer Dirichlet Process and Beta Process individually, as the single valued $\alpha$ parameter is replaced by a set of individually-different valued $\{ \prod_f \eta_{fk}^{\phi_{if}} \}$.

\section{Modelling Link Data \label{sec_link_data}}
As stated in the introduction, many real-world applications use directional count and unit link data instead of binary link data. Thus, we need more appropriate generation distributions to model them. We here mainly discuss the {MMSB} case, the {LFRM} case and its detailed derivations is included in the Supplementary Material.

%The binary link data setting is a default assumption for all the previous relational models, where $1$ for the existence of a relation and $0$ vanishes this relation. However, as stated in the introduction, other link data exist in the real world application, including count and unit link data. As these formats contain discriminative information on the link data, manually ``binarizing'' them could not fully utilize them.

\subsection{Count Link Data}
%Rather than a binary setting of relational data in previous works, a more informative representation of the relational data
%the count relational data, which is common in real world social network dataset.
%The Poisson Distribution parameterized by the corresponding role-compatibility value $B$ is used here to model the count link data.
%Count link data describes more details on the granular of a relation. In here, we try to assign this count link data problem by using Poisson distribution as the generation distribution, instead of a Bernoulli distribution previously.
In iMMM, we propose to model the count link data with the following likelihood and prior distributions:
\begin{equation}
\begin{aligned}
e_{ij}\sim \textrm{Poisson}(B_{s_{ij}r_{ij}})\\
B_{kl}\sim \textrm{Gamma}(\alpha_{B}, \beta_{B})\\
\end{aligned}
\end{equation}
The parameter $B_{kl}$ in the Poisson distribution reflects the compatibility between the two communities $k$ and $l$. The lager $B_{kl}$ encourages larger $e_{ij}$. Further, we put a vague gamma prior on it as $B_{kl}\in R^+$. The resultant predictive distribution is the Negative Binomial distribution \cite{hilbe2011negative} when marginalizing out $B_{kl}$

For the discovered communities $k,l$ in iMMM, we sample $B_{kl}$ as:
\begin{equation}
\begin{split}
& B_{kl}\sim \textrm{Gamma}(\sum_{i'j'}e_{i'j'}+\alpha_B, m_{kl}+\beta_B) \\
\end{split}
\end{equation}
Here $i'j' = \{ij|s_{ij}=k, r_{ij}=l\}, m_{kl}=\#\{i'j'\}$.

On calculating an undiscovered community $K+1$ in iMMM, we can marginalize the corresponding $B$ value out:
\begin{equation} \label{eq_2}
\Pr(e_{ij}|\alpha_B, \beta_B)=\frac{\beta_B^{\alpha_B}}{e_{ij}!\cdot(\beta_B+1)^{\alpha_B+e_{ij}}}\cdot\frac{\prod_{q=0}^{e_{ij}}(\alpha_B+q)}{\alpha_B+e_{ij}}
\end{equation}

\subsection{Unit Link Data}
%Another interesting case is dealing with the unit link data. The unit link data lies in the unit interval $[0, 1]$, examples using this link data type includes the likes or dislikes based on one standard, the proportion of point agreement between two persons. A beta distribution on this unified link data is a natural option chosen as the generation distribution.

In modelling the unit Link Data, we use the Beta Distribution instead as the corresponding generation distribution. For the {iMMM} case, we have:
\begin{equation}
\begin{aligned}
e_{ij}\sim \textrm{Beta}(B_{s_{ij}r_{ij}}, 1)\\
B_{kl}\sim \textrm{Gamma}(\alpha_B, \beta_B)\\
\end{aligned}
\end{equation}
Thus, the  posterior distribution of $B_{kl}$ between the discovered communities $k,l$ is:
%\begin{equation}
%\begin{split}
%& p(B_{kl}|z, \alpha_B, \beta_B)\propto \prod_{ij:s_{ij}=k, r_{ij}=l}p(e_{ij}|B_{kl})p(B_{kl}|\alpha_B, \beta_B)\\
%\propto & B_{kl}^{m_{kl}}\left(\prod_{i_1j_1}e_{ij}\right)^{B_{kl}}\cdot B_{kl}^{\alpha_B-1}e^{-\beta_BB_{kl}}\\
%\propto & B_{kl}^{m_{kl}+\alpha_B-1}e^{-\left(\beta_B-\sum_{i_1j_1}\ln e_{ij}\right)B_{kl}} \\
%\Longrightarrow & B_{kl}\sim Gamma(m_{kl}+\alpha_B, \beta_B-\sum_{i_1j_1}\ln e_{ij}).
%\end{split}
%\end{equation}
\begin{equation}
\begin{split}
& B_{kl}\sim \textrm{Gamma}(m_{kl}+\alpha_B, \beta_B-\sum_{i_1j_1}\ln e_{ij})\\
\end{split}
\end{equation}

While involving an undiscovered community $K+1$, we have
\begin{equation}
p(e_{ij}|\alpha_B, \beta_B)=\frac{\alpha_B}{e_{ij}}\cdot\frac{\beta_B^{\alpha_B}}{(\beta_B-\ln e_{ij})^{\alpha_B+1}}
\end{equation}

%For LFRM,
%\begin{equation}
%\begin{aligned}
%B_{kl}\sim Gamma(\alpha_B, \beta_B)\\
%e_{ij}\sim Beta(\prod_{k,l}B_{kl}^{z_{ik}z_{jl}}, 1)\\
%\end{aligned}
%\end{equation}
%Thus, we get
%\begin{equation}
%\begin{split}
%& p(B_{kl}|z, \alpha_B, \beta_B)\propto \prod_{ij:s_{ij}=k, r_{ij}=l}p(e_{ij}|B_{kl})p(B_{kl}|\alpha_B, \beta_B)\\
%\propto & B_{kl}^{m_{kl}}\left(\prod_{i_1j_1}e_{ij}\right)^{\prod_{kl}B_{kl}^{z_{ik}z_{jl}}}\cdot B_{kl}^{\alpha_B-1}e^{-\beta_BB_{kl}}\\
%\propto & B_{kl}^{m_{kl}+\alpha_B-1}e^{-\left(\beta_B-\sum_{i_1j_1}\prod_{k_1l_1\neq kl}B_{k_1l_1}^{z_{ik_1}z_{jl_1}}\ln e_{ij}\right)B_{kl}} \\
%\Longrightarrow & B_{kl}\sim Gamma(m_{kl}+\alpha_B, \beta_B-\sum_{i_1j_1}\prod_{k_1l_1\neq kl}B_{k_1l_1}^{z_{ik_1}z_{jl_1}}\ln e_{ij}).
%\end{split}
%\end{equation}
%Here $i_1j_1 = \{ij|z_{ik}=1, z_{jl}=1\}, m_{kl}=\#\{i_1j_1\}$.

%\subsection{Notations}
%As the binary format has been extensively studied in previous works, we treat it as an grountruth method.
%
%Here we should note that although there seems no technical difficulties in modelling these formats of link data, other than the binary setting, the generality and adaption for the real world application makes it an emergent topic needs to be assigned.

\section{Inference \label{sec_4}}

\subsection{Without Collapsing $\pi_i$}
Due to the space limit, the detailed sampling procedures of both InfMM and InfLF models are summarised in the Supplementary Material.  In here, we explicitly sample $\{\pi_i\}_{i=1}^n$. In fact, most of the conditional distributions used in Gibbs have been stated in the preceding sections.

\begin{table}
\caption{Computational Complexity for Different Models} \label{table_0}
\centering
\begin{tabular}{c|c}
\hline
Models & Computational complexity \\
  \hline
{IRM}  &  $\mathcal{O}(K^2n)$ \cite{PalKnoGha12}\\
{LFRM}  &  $\mathcal{O}(K^2n^2)$ \cite{PalKnoGha12}\\
{MMSB}  &  $\mathcal{O}(Kn^2)$ \cite{conf/icml/KimHS12}\\
{NMDR} &  $\mathcal{O}(Kn^2+Kn+FKn)$ \\
{InfMM}  &  $\mathcal{O}(Kn^2+Kn+FKn)$ \\
{InfLF}  &  $\mathcal{O}(K^2n^2+Kn+FKn)$ \\
  \hline
  \end{tabular}
\end{table}

\subsection{$\pi_i$-Collapsed Sampling for InfMM}
We have further improved the sampling strategy by collapsing the membership distribution $\pi_i$ for the finite communities case of {InfMM}, in which we have analysed its computational complexity.
Under the condition of finite communities number, inferencing the InfMM by collapsing the mixed-membership distributions $\{\pi_i\}_i^{n}$ is a promising solution. W.l.o.g., the membership indicators' joint probability for entity $i$ is:
\begin{equation}
\Pr(\{s_{ij}\}_{j=1}^n,\{r_{ji}\}_{j=1}^n|\phi, \eta)\propto \frac{\prod_k\Gamma(\Gamma(N_{ik}+\prod_f \eta_{fk}^{\phi_{if}}))}{\Gamma(2n+\sum_k\prod_f \eta_{fk}^{\phi_{if}})}
\end{equation}
Thus, the conditional probability of the membership indicator $s_{ij}$ (or $r_{ij}$) is:
\begin{equation}
\Pr(s_{ij}=k|z_{\backslash s_{ij}}, \phi, \eta)\propto N_{ik}^{\backslash s_{ij}}+\prod_f \eta_{fk}^{\phi_{if}}
\end{equation}
Comparing to its counterpart in MMSB \cite{airoldi2008mixed}:
\begin{equation} \label{eq_3}
\Pr(s_{ij}=k|z_{\backslash s_{ij}}, \phi, \eta)\propto N_{ik}^{\backslash s_{ij}}+\frac{\alpha}{K}
\end{equation}
our \emph{collapsed InfMM} (cInfMM) replace the term $\frac{\alpha}{K}$ in Eq. \ref{eq_3} with $\prod_f\eta_{fk}^{\phi_{if}}$. In fact, while the MMSB generates the membership distribution $\pi_i$ through the Dirichlet distribution with parameters $(\frac{\alpha}{K}, \cdots, \frac{\alpha}{K})$, our cInfMM's parameter is vector containing unequal elements $(\prod_f\eta_{f1}^{\phi_{if}}, \cdots, \prod_f\eta_{fK}^{\phi_{if}})$.

%The {MMSB} generates $\pi_i$ through $Dirichlet(\frac{\alpha}{K}, \cdots, \frac{\alpha}{K})$, while our collapsed {InfMM} is $Dirichlet(\prod_f\eta_{f1}^{\phi_{if}}, \cdots, \prod_f\eta_{fK}^{\phi_{if}})$.

Due to the unknown information on the undiscovered communities, we are limiting our cInfMM model into this finite communities number case. The extension on the infinite communities case remains a future task.

\subsubsection{Computational Complexity}
We estimate the computational complexities for each model and present the results in Table \ref{table_0}. Our InfMM and InfLF are $\mathcal{O}(Kn^2+Kn+FKn)$ and $\mathcal{O}(K^2n^2+Kn+FKn)$ respectively, with $\mathcal{O}(Kn)$ for the sampling of $\{\pi_i\}_{i=1}^n$ and $\mathcal{O}(FKn)$ for the metadata information's incorporation.
%the likelihood's computational load is also $\mathcal{O}(K^2N^2)$, since the latent variables $s_{ij}$ and $r_{ij}$ needs to be sampled together for each link data $e_{ij}$, which would need a $K\times K$ sampling table. For the importance indicator matrix $\eta$, $\mathcal{O}(F^2K)$ and $\mathcal{O}(KN)$ calculations are needed both for the {logistic normal} and {exponential-sum-logarithm} operation. Lastly, the role-compatibility matrix's computational cost is $\mathcal{O}(K^2N)$ as $N$ for allocating the latent features. Thus, a total of $\mathcal{O}(K^2N^2+F^2K+KN+K^2N)$ calculations is needed for one iteration in all of the models.

%\begin{figure*}[htbp]
%\centering
%\includegraphics[scale=0.80, width = 0.80 \textwidth, bb = 103 308 515 511, clip]{images/syn_images.eps}
%\caption{Posterior predictive distributions for different models. {\color{red}There will be some revise and comments.}}
%\label{fig_6}
%\end{figure*}
\section{Experiments \label{sec_5}}
We analyse the performance of our models (InfMM and InfLF) on three real-world datasets: lazega-lawfirm dataset \citep{lazega2001collegial}, MIT Reality Mining dataset \cite{Eagle2006RMS} and NIPS Co-authoring dataset \cite{TehGor2009a}. The Previous works for comparison including IRM \cite{kemp2006learning}, LFRM \cite{miller2009nonparametric}, iMMM \cite{koutsourelakis2008finding} (an infinite community case of MMSB \cite{airoldi2008mixed}), and NMDR \cite{conf/icml/KimHS12}  are brought here to validate our framework's behaviour.
%, {infinite latent attribute model (ILA)} \cite{PalKnoGha12}.

We have independently implemented the above benchmark algorithms to the best of our understanding. There have been a slight variation to NMDR, in which we have employed Gibbs sampling to sample the unknown cluster number, instead of Retrospective MCMC \cite{papaspiliopoulos2008retrospective} used in the original paper. This is because we have set the conjugate priors to their corresponding generation distributions.

To validate the model's prediction accuracy, we use a ten-fold cross-validation, where we randomly select one out of ten for each entity's link data as testing data and the rest as training data. The criteria for evaluating the prediction capability are the training error ($0-1$ loss), the testing error ($0-1$ loss), the testing log likelihood and the AUC (Area Under the roc Curve) score, where these detailed derivations can be found in the Supplementary Material. Also, an extra study is performed on learning the importance indicator of metadata information in the lazega-lawfirm dataset as we have successfully inferred the corresponding $\eta$ values.

At the beginning of the learning process, we set the vague Gamma prior $\Gamma(1, 1)$ for all the hyper-parameters, including $\alpha_{\eta}, \beta_{\eta}, \alpha_B, \beta_B$. The initial states are of random guesses on the hidden labels (membership indicators in MMSB and latent feature in LFRM). For all the experiments, we run chains of $10,000$ MCMC samples for 30 times, assuming $5000$ samples are used for burn-in. The average of the remaining $5000$ samples are reported.

%Figure \ref{fig_6}. shows the detail outcome of recovered images by different models.

\begin{table*} [htbp]
\centering
\caption{Performance on Real world Datasets (Mean $\mp$ Standard Deviation)} \label{table_2}
\begin{threeparttable}
\begin{tabular}{c|c|cccc}
\hline
Datasets & Models & Training error & Testing error & {\small Testing log likelihood} & AUC \\
\hline
\multirow{7}{*}{Lazega}&{IRM} & $    0.0987 \mp     0.0003$ & $    0.1046 \mp     0.0012$ & $ -201.7912 \mp     3.3500$
& $    0.7056 \mp     0.0167$ \\
&{LFRM} & $    0.0566 \mp     0.0024$ & $    0.1051 \mp     0.0064$ & $ -222.5924 \mp    16.1985$ & $    0.8170 \mp     0.0197$ \\
&iMMM & $    0.0487 \mp     0.0068$   & $    0.1096 \mp     0.0026$  &  $ -202.7148 \mp     5.3076$   & $    0.8074 \mp     0.0141$  \\
&{NMDR} & $    0.0640 \mp     0.0055$  & $    0.1133 \mp     0.0018$ &  $ -207.7188 \mp     3.4754$   & $    0.8285 \mp     0.0114$ \\
&{InfMM} & $ \boldsymbol{ 0.0334 \mp     0.0056}$  & $\boldsymbol{ 0.1067 \mp     0.0021}$  &  $\boldsymbol{-196.0503 \mp     4.3962}$   &$\boldsymbol{0.8369 \mp     0.0122}$  \\
&{InfLF} & $    0.0389 \mp     0.0126$  & $    0.1012 \mp     0.0034$ &  $ -213.5246 \mp     12.3249$   & $    0.8123 \mp     0.0135$  \\
&{cInfMM}\tnote{1} & $    0.0466 \mp     0.0092$  & $    0.1119 \mp     0.0020$ &  $ -205.0673 \mp     4.5321$   & $    0.8314 \mp     0.0119$  \\
\hline
\multirow{9}{*}{Reality}&{IRM} &  $    0.0627 \mp     0.0002$ & $    0.0665 \mp     0.0004$ & $ -133.8037 \mp     1.1269$ & $    0.8261 \mp     0.0047$ \\
&{LFRM} & $    0.0397 \mp     0.0017$ & $    0.0629 \mp     0.0037$ & $ -143.6067 \mp    10.0592$ & $    0.8529 \mp     0.0179$ \\
&iMMM & $    0.0297 \mp     0.0055$  & $    0.0625 \mp     0.0015$  &  $ -126.7876 \mp     3.4774$   & $    0.8617 \mp     0.0124$  \\
&{NMDR} & $    0.0386 \mp     0.0040$  & $    0.0668 \mp     0.0013$  &  $ -139.5227 \mp     2.9371$  & $    0.8569 \mp     0.0138$  \\
&{InfMM} & $    0.0269 \mp     0.0047$  & $    0.0621 \mp     0.0015$ &  $ -127.7377 \mp     3.1313$   & $    0.8507 \mp     0.0134$  \\
&{InfLF} & $    0.0379 \mp     0.0046$  & $    0.0732 \mp     0.0049$ &  $ -131.0326 \mp     9.4521$   & $    0.8645 \mp     0.0139$  \\
&{cInfMM}\tnote{1} & $    0.0553 \mp     0.0023$  & $    0.0641 \mp     0.0011$  &  $ -126.9091 \mp     2.6459$  & $    0.8597 \mp     0.0099$  \\
&{pInfMM}\tnote{2} & $  \boldsymbol{   0.0212 \mp     0.0012}$  & $  \boldsymbol{   0.0601 \mp     0.0012}$  &  $\boldsymbol{  -121.8172 \mp     4.8434}$  & $  \boldsymbol{   0.8736 \mp     0.0107}$  \\
&{piMMM}\tnote{2} &$    0.0223 \mp     0.0011$  & $    0.0632 \mp     0.0013$  & $ -131.8161 \mp     5.6687$ &  $    0.8631 \mp     0.0098$  \\
\hline
&  {IRM} & $    0.0317 \mp     0.0004$ & $    0.0423 \mp     0.0014$ & $ -135.0467 \mp     7.3816$ & $    0.8901 \mp     0.0162$ \\
NIPS&  {LFRM} &$    0.0482 \mp     0.0794$ & $    0.0239 \mp     0.0735$ & $ -105.2166 \mp   179.5505$ & $    0.9348 \mp     0.0167$ \\
Coauthor & iMMM &$  \boldsymbol{0.0061 \mp 0.0019}$ & $    0.0253 \mp     0.0035$ & $  -83.4264 \mp     9.4293$ & $    0.9574 \mp     0.0155$ \\
&{piMMM}\tnote{2} &$    0.0097 \mp     0.0048$  & $  \boldsymbol{  0.0213 \mp     0.0022}$  & $\boldsymbol{-81.49 \mp 7.9465}$ &  $  \boldsymbol{0.9643 \mp     0.0177}$  \\
\hline
\end{tabular}
      \begin{tablenotes}
        \footnotesize
        \item[1] cInfMM is used to denote the $\pi_i$-\emph{collapsed InfMM};
        \item[2] piMMM and pInfMM are used to denote the models of using Poission distribution as the generation distribution.
      \end{tablenotes}
\end{threeparttable}
\end{table*}

%\begin{table*}[http]
%\caption{Link Prediction performance} \label{table_3}
%\centering
%\begin{tabular}{c|c|cccccccc}
%  \hline
%Dataset & Criterion & {IRM} & {LFRM} & {MMSB} &  iMMM & {NMDR} & {LN-LFRM} & {Beta-MMSB} & {Beta-LFRM}\\
%  \hline
%\multirow{4}{*}{Lazega}&  AUC  &   & &  &  &   &  & &\\
%&  S.D. &   & &  &  &   &  & &\\
%&  TLL &   & &  &  &   &  & &\\
%&  S.D. &    & &  &  &   &  & &\\
%  \hline
%\multirow{4}{*}{MIT Reality}&  AUC  &   & &  &  &   &  & &\\
%&  S.D. &   & &  &  &   &  & &\\
%&  TLL &   & &  &  &   &  & &\\
%&  S.D. &    & &  &  &   &  & &\\
%\hline
%\end{tabular}
%\end{table*}

\subsection{Lazega-lawfirm Dataset}
The lazega-lawfirm dataset is about a social networking corporate located in the northeastern part of the U.S. in 1988 - 1991. The dataset contains three different types of relations: co-work network, basic advice network, and friendship network for 71 attorneys, in which each link data is labelled as $1$ (exist) or $0$ (absent). Apart from these three $71\time 71$ binary asymmetric matrices, the datasets also provide some metadata information on each of the attorneys, including the status (partner or associate), gender, office (Boston, Hartford or Providence), years with the firm, age, practice (litigation or corporate), law school (harvard, yale, ucon or other). After binarizing these attributes, a $71\times 11$ binary metadata information matrix is obtained.

We conduct the link prediction on the co-work network and show the result in Table \ref{table_2}. Notably, the performance of our implementation of NMDR model is inferior compared to its original \cite{conf/icml/KimHS12}. The reason may be as a result of a sub-optimal metadata binarization process.  However, we have shown that with the same attributes, our InfMM performs better than the NMDR in terms of training error, test capability and convergence behaviour (including the burn in samples needed and mixing rate).

%
%\begin{figure}[htbp]
%\centering
%\includegraphics[scale=0.45, width = 0.45 \textwidth]{images/trace_plot_lazega.eps}
%\caption{Graphical model of NLFRM Incorporating Metadata Information. As we can see, the model {ESL-MMSB} is the first to reach a stable status of the Markov chain and its performance is also the best among all the others. The {NMDR} model is the slowest to reach the stable status and its performance is also not encouraging. The {Collapsed-ESL-MMSB} performs quite similar as the {ESL-iMMM}, despite the a simpler implementation it needs.}
%\label{fig_5}
%\end{figure}

Another interesting topic here is the learning of the importance indicator $\eta$ for the attributes in the metadata information. We take a geometric mean value of participating the communities for each attribute and show the detail result in Table \ref{table_4}. As stated in Section \ref{sec_31}, smaller value indicates larger influence.
\begin{table} [htbp]
\centering
\caption{The recovered Importance indicator $\eta$ in Lazega-lawfirm dataset. As we can see, the value of attribute \emph{office} is the smallest amongst all these values, which indicates it has a largest influence. This phenomenon is consistent with common sense as people in the same place would be more likely to have a relation. Also, the attribute \emph{practice} is more important than the others in forming this co-work link data.} \label{table_4}
\begin{tabular}{cc|cc}
\hline
\multirow{2}{*}{office} & $\boldsymbol{0.5596}$ & \multirow{2}{*}{age} &  1.0802 \\
 & $\boldsymbol{ 0.4996}$ & &   0.8114 \\
\hline
{\small years with}  & $\boldsymbol{0.5258}$ &  \multirow{2}{*}{law school}  &  0.8268 \\
{\small the firm}  &  0.8207 & &  1.1061 \\
\hline
status &  0.8371& practice  &$\boldsymbol{ 0.5585}$  \\
\hline
gender  & 0.7307& &\\
\cline{1-2}
\end{tabular}
\end{table}

\subsection{MIT Reality Mining}
Based on the MIT Reality Mining dataset, we obtain a proximity matrix describing each entity's proximity towards the others, i.e., $e_{i,j}$ represents the proximity from $i$ to $j$ based on participant $i$'s opinion. The detailed link values indicate the average proximity from one subject to another, which is categorised into 5 values correspondingly. While using previous models \cite{koutsourelakis2008finding}, we manually set the proximity value larger than 10 minutes per day as 1, and 0 otherwise. We hence obtain a $73\times 73$ asymmetric matrix. According to the generation distribution used, we can choose either the integer matrix or its binary version.

Alongside this directional link data, we also have a survey data on the entities involving metadata, including the traffic choice to work, personal habitat, social activity, the communication method, and satisfaction of university life.

%The dataset are roughly divided into four groups: Sloan Business School students (Sloan), Lab faculty, senior students with more than 1 year in the lab and junior students.
%
%In the count link data test, we test both cases of the binary link data and count link data. The binary version is set as the benchmark for the link prediction.

As we can see in Table \ref{table_2}, we find our InfMM's performance is similar to the ones in iMMM. The reason may be that the metadata information does not correlate with the link data. Our pInfMM and piMMM's performance is the best among all these models. This validates the necessity of using Poisson distribution while encountering the count link data.

%\begin{figure}[htbp]
%\centering
%\includegraphics[scale=0.45, width = 0.45 \textwidth]{images/trace_plot_lazega.eps}
%\caption{Graphical model of NLFRM Incorporating Metadata Information. As we can see, the model {ESL-MMSB} is the first to reach a stable status of the Markov chain and its performance is also the best among all the others. The {NMDR} model is the slowest to reach the stable status and its performance is also not encouraging. The {Collapsed-ESL-MMSB} performs quite similar as the {ESL-iMMM}, despite the a simpler implementation it needs.}
%\label{fig_2}
%\end{figure}
\subsection{Convergence Behaviour \label{sec_61}}

\begin{table*} [htbp]
\centering
\caption{Mixing rate (Mean $\mp$ Standard Deviation) for different models, with the bold type denoting the best ones within each row. As we can see, our model InfMM performs the best among all the models. } \label{table_1}
\begin{tabular}{cc|ccccc}
\hline
Datasets & Criteria & iMMM & LFRM & NMDR  & InfMM & InfLF\\
\hline
%\multirow{2}{*}{Synthetic}&$\hat{\tau}$ &  &  &  &    \\
%&ESS &  &  &  &      \\
%\hline
\multirow{2}{*}{Lazega}&$\hat{\tau}$ & $     166.2 \mp      90.37$  & $     310.6 \mp     141.95$   & $     179.8 \mp     156.96$ & $   \boldsymbol{   39.1 \mp      40.58}$ & $     149.2 \mp     126.12$   \\
&ESS & $      77.6 \mp      38.71$ & $      40.7 \mp      26.26$  & $     134.3 \mp     133.12$  & $  \boldsymbol{   341.8 \mp     132.00}$ & $61.2 \mp 59.93$\\
\hline
\multirow{2}{*}{Reality}&$\hat{\tau}$ &$     184.9 \mp      78.88$  & $     113.4 \mp      77.35$  & $     142.8 \mp     129.99$   & $ \boldsymbol{ 27.8 \mp  22.49}$ & $134.2 \mp 163.23$\\
&ESS  & $      62.5 \mp      22.70$ & $     125.5 \mp      71.93$  & $     185.0 \mp     206.12$ & $ \boldsymbol{ 449.7 \mp  181.37}$ & $71.24\mp 48.74$\\
\hline
\end{tabular}
\end{table*}

\begin{figure*}[htbp]
\begin{minipage}[b]{0.45\linewidth}
\centering
\includegraphics[width=\textwidth]{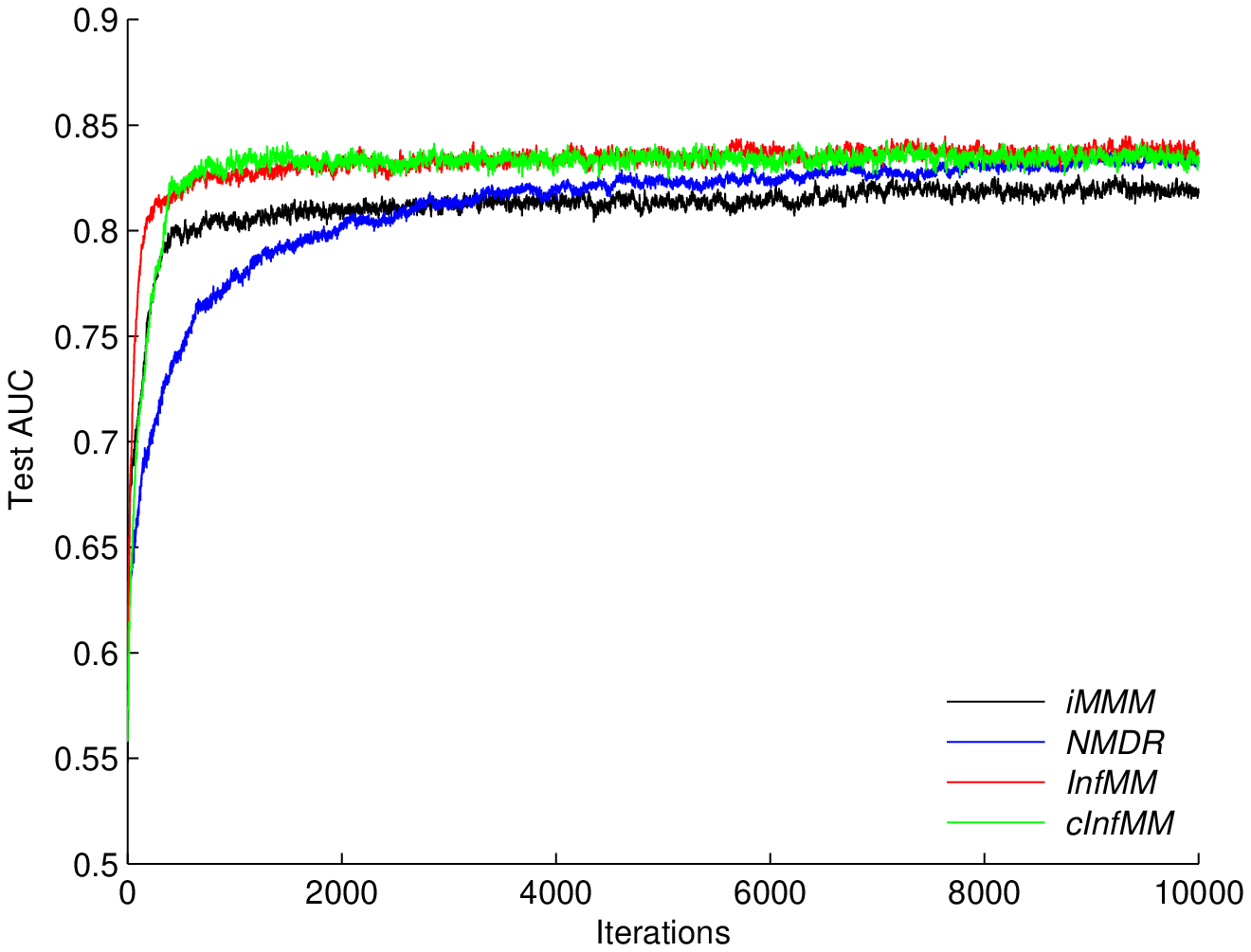}
\end{minipage}
\hspace{0.5cm}
\begin{minipage}[b]{0.45\linewidth}
\centering
\includegraphics[width=\textwidth]{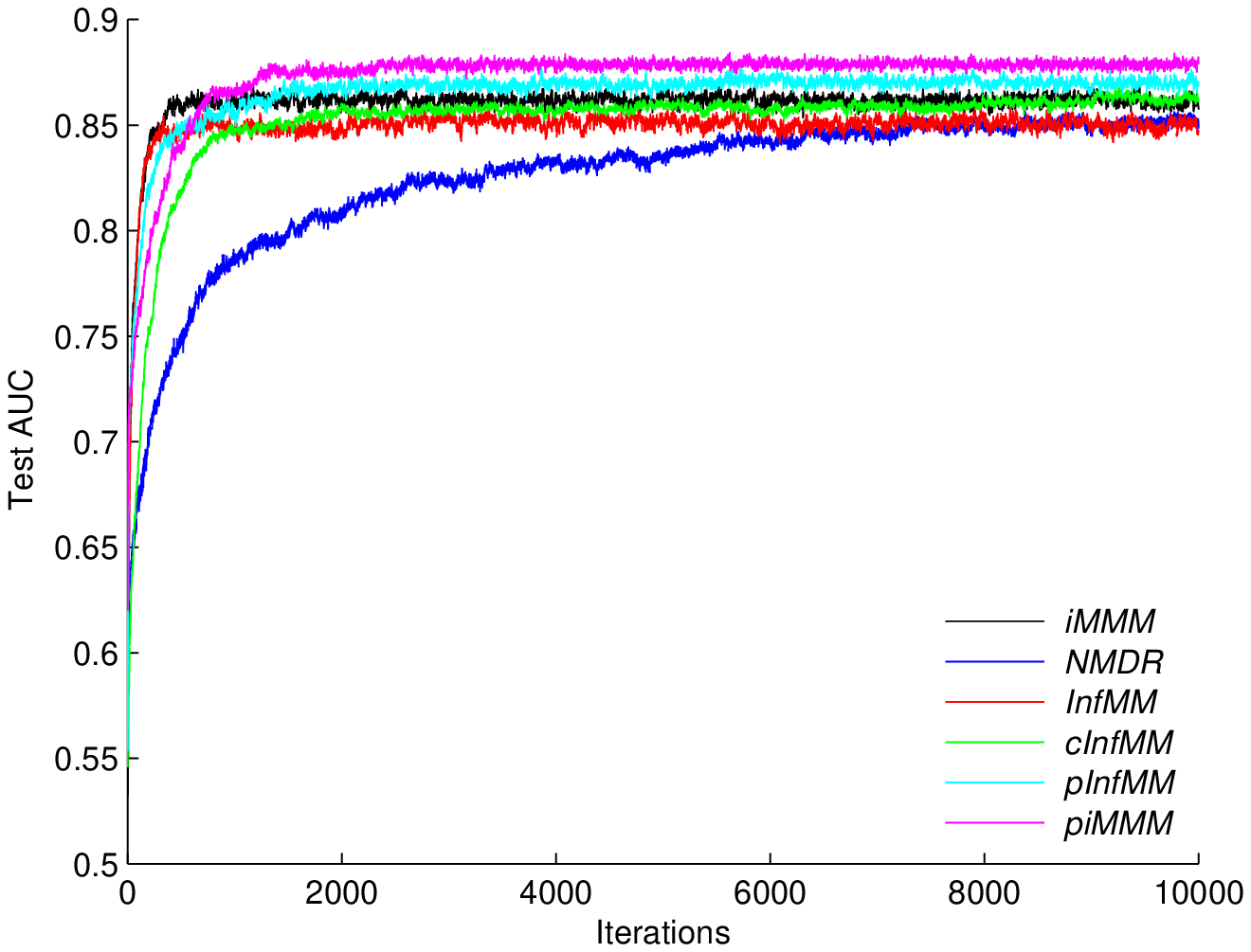}
\end{minipage}
\caption{Left: Lazega dataset; Right: MIT Realtiy dataset. Trace plot of the AUC value versus iteration time in different MMSB type models. As we can see, except for NMDR, all the other models reach the stable status quite fast. On the lazega dataset, our InfMM and cInfMM outperform all the others. On the MIT Reality dataset, our InfMM and cInfMM do not perform better than iMMM. However, our pInfMM and piMMM perform the best compared to other models.}
\label{fig_2}
\end{figure*}
\paragraph{Trace plot for AUC}
A trace plot for the AUC value versus iteration time could help us choose an appropriate burn-in length. An earlier reach to the stable status of MCMC is desirable as it indicates fast convergence. Figure \ref{fig_2} shows the detailed results.
%\begin{figure}
%\centering
%\includegraphics[scale=0.45, width = 0.45 \textwidth]{images/trace_plot.eps}
%\caption{Trace plot of the training error.{\color{red}Add some comments.}}
%\label{fig_4}
%\end{figure}
%\begin{table*} [htbp]
%\centering
%\caption{Iteration times} \label{table_2}
%\begin{tabular}{c|ccccccc}
%\hline
%Models & {IRM} & {MMSB} & {LFRM} & {NMDR} & {LN-LFRM} & {Beta-MMSB} & {Beta-LFRM} \\
%\hline
%Iteration time &  &  &  &  &  &  &  \\
%S.D. &  &  &  &  &  &  &  \\
%\hline
%\end{tabular}
%\end{table*}

\paragraph{Mixing rate for a stable MCMC.}
Besides the MCMC trace plot, another interesting observation is the mixing rate of the stable MCMC chains. We use the number of active communities $K$ as a function of the updated variable to monitor the mixing rate of the MCMC samples, whereas the efficiency of the algorithms can be measured by estimating the integrated autocorrelation time $\tau$ and Effective Sample Size (ESS) for $K$. $\tau$ is a good performance indicator as it measures the statistical error of Monte Carlo approximation on a target function $f$. The smaller $\tau$, the more efficient of the algorithm. Also, the ESS of the stable MCMC chains informs the quality of the Markov chains, i.e., a larger ESS value indicates more independent useful samples, which is our desired property.

On estimating the integrated autocorrelation time, different approaches are proposed in \cite{geyer1992practical}. Here we use an estimator $\widehat{\tau}$ \cite{kalli2011slice} and the ESS value is calculated based on $\widehat{\tau}$ as:
\begin{equation}
\widehat{\tau}=\frac{1}{2}+\sum_{l=1}^{C-1}\widehat{\rho}_l;\; ESS= \frac{2M}{1+\widehat{\tau}}.
\end{equation}
Here $\widehat{\rho}_l$ is the estimated autocorrelation at lag $l$ and $C$ is a cut-off point which is defined as $C:=\min\{l:|\widehat{\rho}_l|<2/\sqrt{M}\}$, and $M$ equals to half of the original sample size, as the first half is treated as a burn in phase. The detailed results are shown in Table \ref{table_1}.

\subsection{NIPS Coauthor Dataset}
We use the co-authorship as a relation gained from the proceedings of the Neural Information Processing Systems (NIPS) conference for years 2000-2012. Due to the sparsity of the co-authorship, we observe the author activities in all 13 years (i.e. regardless of the time factor) and set the link data being 1 if two corresponding authors have co-authored on no less than 2 papers, which is to remove the co-authoring randomness. Further, the authors with less than 4 relationships with others are manually eliminated. Thus, a $92\times92$ symmetric, binary matrix is obtained.

We focus on the count link data's modelling in this dataset, where the actual link data among these 92 entities are used. As the detail result shown in Table \ref{table_2}, our piMMM performs better than the classical iMMM.

%One interesting part of this dataset is that the diagonal value of the relational matrix is 1 for all the selected authors. This fact is quite intuitive as an author always `co-author' with himself or herself on every paper.

\section{Conclusions \& Future work\label{sec_6}}
Increasing applications with natural and social networking behaviors request the effective modelling of hidden relations and structures. This is beyond the currently available models, which only involves limited link information in binary settings.
In this paper, we have proposed a unified approach to incorporate various kinds of information into the relational models, including the metadata information and different formats of link data. The proposed \emph{informative mixed membership} (InfMM) model and \emph{informative latent feature} (InfLF) model have been demonstrated effective in learning the structure and show advanced performance on learning implicit relations and structures. Also, our adaptive link data modelling method further boosts the capability of utilizing rich link data in the real- work scenario.

We are extending our work to: 1), how to integrate the multi-relational networks and unify them into the InfMM framework to deeply understand network structure; 2), when the link data is categorical, what will be a proper generation distribution to describe the linkage; 3), as there are more advanced constructions for the beta process \cite{paisley2010stick,paisley2012stick}, what are more flexible ways to incorporate the metadata information into LFRM; and 4), when the metadata information goes beyond the binary scope and becomes the continuous form, we need an effectively way to utilize such information.

{\small
\bibliography{LFRMIMI}
\bibliographystyle{icml2014}
}
\end{document}